\documentclass[10pt,twocolumn,letterpaper]{article}

\usepackage{adjustbox}
\usepackage[pagenumbers]{cvpr} 

\usepackage{graphicx}
\usepackage{amsmath}
\usepackage{amssymb}
\usepackage{booktabs}

\usepackage[pagebackref,breaklinks,colorlinks]{hyperref}

\usepackage[capitalize]{cleveref}
\crefname{section}{Sec.}{Secs.}
\Crefname{section}{Section}{Sections}
\Crefname{table}{Table}{Tables}
\crefname{table}{Tab.}{Tabs.}

\def\confName{CVPR}
\def\confYear{2022}

\begin{document}

\title{3D Textured Shape Recovery with Learned Geometric Priors}


\author{
Lei Li$^*$\\
ETH Zurich\\
{\tt\small leilil@ethz.ch}
\and
Zhizheng Liu$^*$\\
ETH Zurich\\
{\tt\small liuzhi@ethz.ch}
\and
Weining Ren$^*$\\
ETH Zurich\\
{\tt\small weiren@ethz.ch}
\and
Liudi Yang$^*$\\
ETH Zurich\\
{\tt\small liudyang@ethz.ch}
\and
Fangjinhua Wang\\
ETH Zurich\\
{\tt\small fangjinhua.wang@inf.ethz.ch}
\and
Marc Pollefeys\\
ETH Zurich\\
{\tt\small marc.pollefeys@inf.ethz.ch}
\and
Songyou Peng\\
ETH Zurich\\
{\tt\small songyou.peng@inf.ethz.ch}
}
\maketitle
{\def\thefootnote{$*$}\footnotetext{~Equal Contributions}}
\begin{abstract}
3D textured shape recovery from partial scans is crucial for many real-world applications. Existing approaches have demonstrated the efficacy of implicit function representation, but they suffer from partial inputs with severe occlusions and varying object types, which greatly hinders their application value in the real world. This technical report presents our approach to address these limitations by incorporating learned geometric priors. To this end, we generate a SMPL model from learned pose prediction and fuse it into the partial input to add prior knowledge of human bodies. We also propose a novel completeness-aware bounding box adaptation for handling different levels of scales and partialness of partial scans. Our code is available at \url{https://github.com/BoSmallEar/sharp2022}.

\end{abstract}

\section{Introduction}\label{introduction}
In this technique report, we demonstrate our solution to the 3rd SHApe Recovery from Partial textured 3D scans (SHARP) challenge, which wins the track 2 of the challenge and ranks 2nd overall. Recovering 3D textured shapes from partial scans is useful and practical in many applications such as augmented reality, but it is a challenging task especially with partial scans obtained from real world. The partial scans can have various sizes and appearances and may suffer from severe occlusion and motion blur. 

Existing approaches on 3D shape recovery have shown promising results with an implicit function representation. For example, IF-Net~\cite{chibane2020implicit} and ConvOcc-Net~\cite{peng2020convolutional} first process the partial scan into spatial features and then representing the completed shape as an implicit occupancy map and the texture as an implicit texture field. While these approaches have great performance in simple recovery tasks, we observe two main failure cases in this year's challenge . 1) They often fail when the partial scan has too many missing parts. When a whole arm of the human is missing, they tend to predict an incomplete arm with many artifacts or nothing at all.  2) They can only handle objects with known sizes and similar shapes. However, in track 2 of the challenge, the objects have varying appearances and sizes. We hypothesize these happen as 1) Given limited training data, it is hard to infer the missing body parts when there is no prior information about the human body. 2) Before obtaining the spatial feature, a bounding box is required to first convert the partial scan to a 3D grid. When the object size is varying, the bounding box often fails to match the completed shape.

To solve these issues, we propose to add geometric priors to help regularize the prediction of completed shapes. 1) For recovering human bodies (track 1), we first predict the human pose and convert it into a SMPL model~\cite{loper2015smpl} as the generic shape to guide the completion of missing body parts.  2) For recovering of universal objects,  we observe that the bounding box acts as a strong prior for the final predicted shape. Therefore, we propose a novel completeness-aware bounding box adaptation module, which learns to adjust the size of the bounding box adaptively to objects of different sizes and partialness. Finally, we further boost the performance by combining the predicted complete shape and the initial partial scan in a shape fusion stage. Thanks to the learned geometric priors, our approach greatly outperforms the baseline implicit-function-based approaches.

\begin{figure*}[ht]
  \begin{center}
  \includegraphics[width=0.95\linewidth]{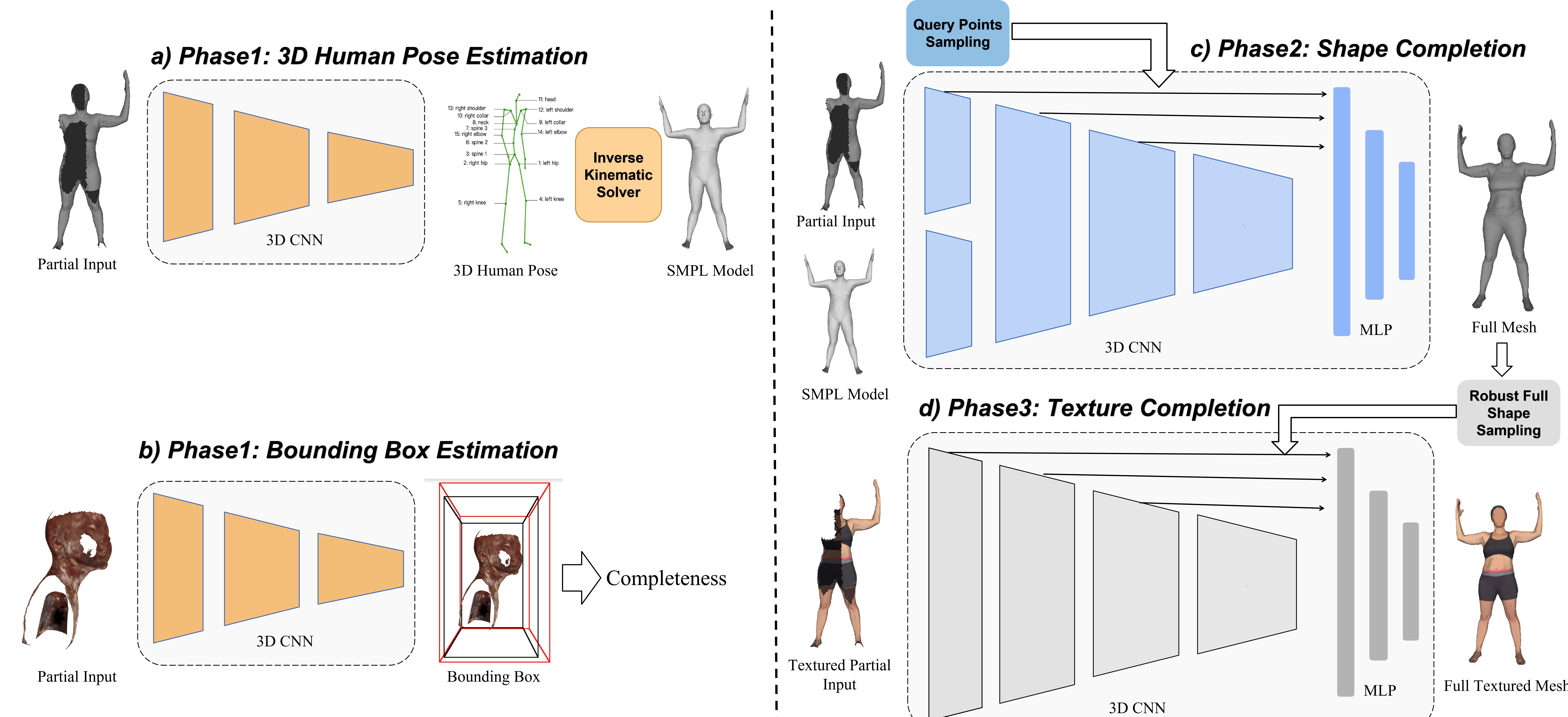}  
  \end{center}

  \caption{Network Architecture. We use three phases to complete our reconstruction. For the geometric prior, we first apply the 3D CNN module to extract the global features and map them to the 3D human pose and bounding box, respectively. For the phase2 of the human body completion task, features from the SMPL model are aggregated with input partial features to provide additional information. In texture completion, the ground truth shape is used as query points for training and the predicted full shape in the phase2 is used as query points for inference.}
  \label{fig:pipeline}
  \vspace{-3mm}
\end{figure*}
\section{Method}

\subsection{Implicit Functions Learning from Surfaces}\label{implicit_functions}
Implicit methods in the 3D area represent the surface as the level-set of a continuous function. 
Implicit function learning shows great strong potential to reconstruct objects, for it can model arbitrary resolution or topology. 
It consists of feature encoding block and feature decoding block (Fig.\ref{fig:pipeline}a and Fig.\ref{fig:pipeline}b).

\noindent\textbf{Encoding from Occupancy Surfaces:} 
The encoder takes voxelized data as input. It is mainly composed of 3D CNN modules and the maxpooling layers followed by instance normalization~\cite{ulyanov2016instance} and ReLU activation. Features are extracted hierarchically with increased receptive fields to obtain local details and global context.
Following the design in IF-Net~\cite{chibane2020implicit}, the multi-scale feature grids $\mathcal{X}_1,\mathcal{X}_2,\cdots,\mathcal{X}_n$ are obtained by utilizing 3D CNN and the maxpooling recursively to downscale the features.
\begin{equation}
\begin{aligned}
\mathcal{X}_{l} &= \operatorname{Maxpolling}(\mathcal{X'}_{l})\\
\mathcal{X'}_{l}&=\operatorname{IN}(\epsilon(\operatorname{3DCNN}(\mathcal{X}_{l-1})))
\end{aligned}
\end{equation}
where IN denotes instance normalization, and $\epsilon$ denotes ReLU activation function. From the encoder, we concatenate multi-scale feature grids to obtain $\mathcal{X}_{0:n}$:
\begin{equation}
\mathcal{X}_{0:n} = [\mathcal{X}_{0}, \mathcal{X}_{1}, \cdots, \mathcal{X}_{n}]
\end{equation}
where $[,]$ denotes the concatenation operator.

\noindent\textbf{Decoding from the Feature Space:}
From the multi-scale features $\mathcal{X}_{0:n}$, we generate the corresponding interpolated features $\mathcal{X}_{0:n}(q)$ with the query points $\mathbf{q} \in \mathbb{R}^{N\times 3}$. The decoder module $\mathcal{F}$ predicts whether there is a surface point at the query point $\mathbf{q}_i$ coordinate.

Typically, neural networks tend to struggle with modeling high-frequency information. High-frequency information in the spatial domain greatly reduces the complexity and frequency the neural network needs to model~\cite{mildenhall2020nerf}. By adding positional embedding in Fourier domain, we can effectively map our features to higher-dimensional space. A shared Multi-layer Perceptron (MLP) is used to decode the features:
\begin{equation}
\mathcal{F}(\mathcal{X}_{0:n}, \mathbf{q}_i)=\operatorname{MLP}([\mathcal{X}_{0:n}(q_i),PE(q_i)])
\end{equation}
where $PE$ denotes the positional embedding.

\subsection{Geometric Priors}\label{geometric_priors}

\noindent\textbf{3D Human Pose Estimation:} The complex human poses and non-rigid deformations make human body reconstruction very challenging, especially when our observations are very vestigial.
Some simple geometric priors, such as the SMPL model~\cite{loper2015smpl}, can guide shape reconstruction with correct 3D human pose and deformation.
The parameters of the SMPL model generally include 3D poses and shapes. We focus only on learning 3D poses, since it is more important to infer the body poses for the vestigial inputs, and reduce the complexity of the geometric prior learning task.
In 3D human pose estimation, a simple skeleton system is usually used to model the vital parts of a human. We describe the human body as a stick model consisting of 25 3D joint coordinates. Following our encoder design, we extract the global feature from the final feature grid $\mathcal{X}_{n}$ with the largest receptive field (Fig.\ref{fig:pipeline}a):

\begin{equation}
\begin{aligned}
\mathcal{G}^p(\mathcal{X}_{n}) = \operatorname{Maxpolling}(\operatorname{MLP}(\mathcal{X}_{n}))
\end{aligned}
\end{equation}
The global feature $\mathcal{G}^p(\mathcal{X}_{n})$ is mapped to the 3D human pose directly. We then optimize the 3D human pose to the SMPL model based on the inverse kinematic solver with basic shape parameters.

\noindent\textbf{Completeness-aware Bounding Box Prediction:} Due to the large-scale variance of universal objects, we cannot use a unified bounding box for all objects like human bodies. We design a bounding box regression network to predict a proper bounding box from partial input. The partial input object is voxelized according to its tight bounding box (coordinate range) $\mathbf{b}_t = [x_t,y_t,z_t,l_t,w_t,h_t]$. The network shares a similar structure with the human pose estimation network:

\begin{equation}
\begin{aligned}
\mathcal{G}^b(\mathcal{X}_{n}) = \operatorname{Maxpolling}(\operatorname{MLP}(\mathcal{X}_{n}))
\end{aligned}
\end{equation}

The global feature $\mathcal{G}^b(\mathcal{X}_{n})$ is mapped to the relative coordinate of 6D bounding box w.r.t the tight bounding box $\tilde{\mathbf{b}}_r = [x_r,y_r,z_r,l_r,w_r,h_r]$, which can be converted to to the absolute coordinate $\mathbf{b}_a = [x_a,y_a,z_a,l_a,w_a,h_a]$ for further shape and texture completion task.
Besides, the size of predicted bounding box can work as a indicator for the incomplete level. We generate the completed mesh only if the bounding box meet either of two criterion:
\begin{align}
     & max\{\mathbf{b}_{a,size}\}/ min\{\mathbf{b}_{a,size}\} > t_1 \\
     & max\{\mathbf{b}_{r,size}\}/ min\{\mathbf{b}_{r,size}\} > t_2
\end{align}
where $\mathbf{b}_{size} = [l,w,h]$ and $t_1, t_2$ are two hyperparameter for completeness.


\subsection{Shape Completion}\label{Shape}
%
%
%
For phase2, to voxelized the input data, we start by normalizing each 3D mesh to a unit box and densely sampling spatial points. Then, we generate occupancy inputs $\mathcal{X}_0 \in \{0,1\}^{K \times K \times K}$ by assigning sampled spatial points to the nearest neighbor grids.
For the query points sampling, we directly sample sufficient points $\mathbf{q}$ near the surfaces to capture more shape details, and convert them into occupancy representations.
%
%
Besides, for the textured human body completion, we extract features from partial inputs and our estimated SMPL models, and concatenate them at an early stage. 
For the decoder, a MLP is used to decode the features into occupancy probabilities:
\begin{equation}
\mathcal{F}^s(\mathcal{X}_{0:n}, \mathbf{q}_i)=\sigma(\operatorname{MLP}([\mathcal{X}_{0:n}(q_i),PE(q_i)]))\mapsto[0,1]
\end{equation}
where $\sigma$ denotes sigmoid activation.
Finally, the predicted occupancy outputs are converted to meshes using marching cubes, and recovered into the original scale. 

Since texture completion is based on shape completion, improving the accuracy of shape completion becomes a critical issue.
%
%
Therefore, we apply our completion results to the missing part only by fusing with the partial inputs to improve the accuracy of shape completion. 
For the scanned regions that exist in original inputs, we retain their original shapes.

\subsection{Texture Completion}\label{Texture} 
%
%
%
The ground truth shape is used for training and the predicted complete shape from phase2 is used for inference.
The encoder takes 4-channel (RGB and occupancy) voxelized data as input.
The step to voxelized inputs is the same as the shape completion. 
For query points, we densely sample on the mesh and add noise in the normal direction to compensate for the error in the shape completion phase.
For the decoder, given a point $p_i$ from the untextured full shape, it is decoded to obtain the color of the surface: 
%
\begin{equation}
\mathcal{F}^t(\mathcal{X}_{0:n}, \mathbf{q}_i)=\operatorname{MLP}([\mathcal{X}_{0:n}(q_i),PE(q_i)])\mapsto[0,255]^3
\end{equation}


\subsection{Loss Functions}
For different phases, we train different networks separately. Therefore, the loss functions of each phase are also independent.

\noindent\textbf{$\ell_{1}$ Loss:} 
%
For 3D human pose estimation, we use the average $\ell_{1}$ loss between the ground truth joint coordinates and the predicted ones: 
%
\begin{equation}
L_{\text {pose }}=\sum_{i=1}^{N_{\text {joint }}}\left\|\hat{t}_{i}-t_{i}\right\|_{1}
\end{equation}
where $\hat{t}_{i}$ is the predicted $i^{th}$ joint coordinate and ${t}_{i}$ is the corresponding ground truth coordinate:
%
%
\begin{equation}
L_{\text {bbox}}=\left\|\hat{\mathbf{b}}-\mathbf{b}\right\|_{1}
\end{equation}
where $\hat{\mathbf{b}}$ is the predicted bounding box parameters and ${\mathbf{b}}$ is the corresponding ground truth one.  
For texture completion, RGB values are directly used to compute loss:
\begin{equation}
L_{\text {texture}}=\frac{1}{N} \sum_{i=1}^{N} \|\hat{\mathbf{c}}_i-\mathbf{c}_i\|_1
\end{equation}
where $\hat{\mathbf{c}}_i$ is the predicted $i^{th}$ RGB value and $\mathbf{c}_i$ is the corresponding ground truth RGB value.  

\noindent\textbf{Balanced BCE Loss:} 
For shape completion, most of the querying points are non-occupied. 
We introduce a balanced BCE loss to solve the imbalance in the querying occupancy.
 
\begin{equation}
L_{\text {shape }}=-\frac{1}{N} \sum_{i=1}^{N} w_{p} \cdot {o}_{i} \cdot \log \left(\hat{o}_{i}\right)+w_{n} \cdot\left(1-{o}_{i}\right) \cdot \log \left(1-\hat{o}_{i}\right)
\end{equation}
where $w_p$ and $w_n$ denote the proportion of voxels with states 1 and 0 in the ground truth, respectively. $\hat{o}_i$ is the predicted $i^{th}$ occupancy probability and ${o}_i$ is the corresponding ground truth occupancy.  
%

\section{Experiments and Results}

\subsection{Implementation Details}
 We implement our network using Pytorch~\cite{paszke2019pytorch}. We take feature grids with $n=6$ scales. We sample 100,000 points for the voxelization preprocessing and the resolution of the grid is $128^3$. We use the Adam optimizer~\cite{kingma2014adam} with learning rate $0.0001$. Our models are trained for 40 epochs on 4 Nivida TITAN Xp GPUs. The batch size on each GPU is set to 1.

\subsection{Recovering Textured Human Body Scans}
\begin{figure}
     \centering
     \begin{subfigure}[t]{0.21\textwidth}
    
         \includegraphics[width=\textwidth]{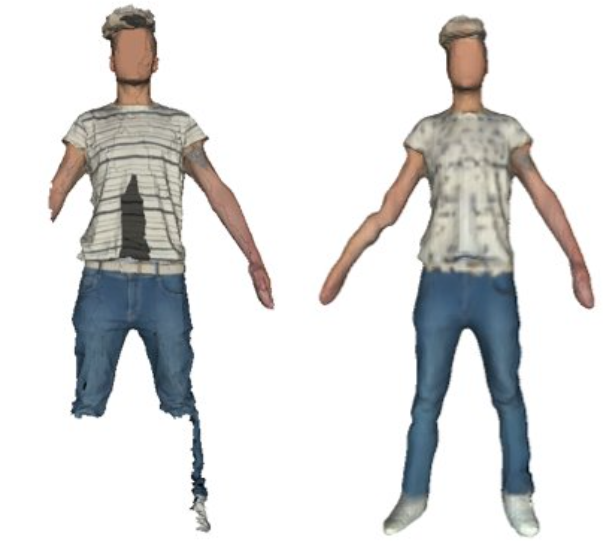}

         \label{fig:track1_1}
     \end{subfigure}
    \medskip
     \begin{subfigure}[t]{0.21\textwidth}

         \includegraphics[width=\textwidth]{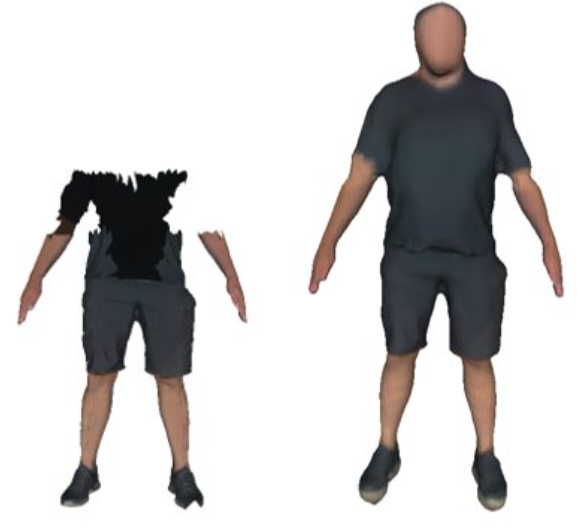}

         \label{fig:track1_2}
     \end{subfigure}

        \caption{Visualization of some results in track 1. The left is partial input and the right is the reconstruction result. Via SMPL model fusion, we can recover error-free human poses for the shape score. And the texture details can also be generated.   }
        \label{fig:track1_result}
        \vspace{-3mm}
\end{figure}

In this track, we train and evaluate our model on the 3DBodyTex.v2 dataset. The challenge is to recover complete textured human body scans from partial bodies of different poses. Considering the potential geometry prior provided by the SMPL model, our model can regenerate the full colored mesh with high accuracy. Some qualitative demos are shown in Fig. \ref{fig:track1_result} , in which the missing parts are replenished appropriately.



\subsection{Recovering Textured Object Scans}
\begin{figure}[htbp]
\centering
     \begin{subfigure}[t]{0.2\textwidth}

         \includegraphics[width=\textwidth]{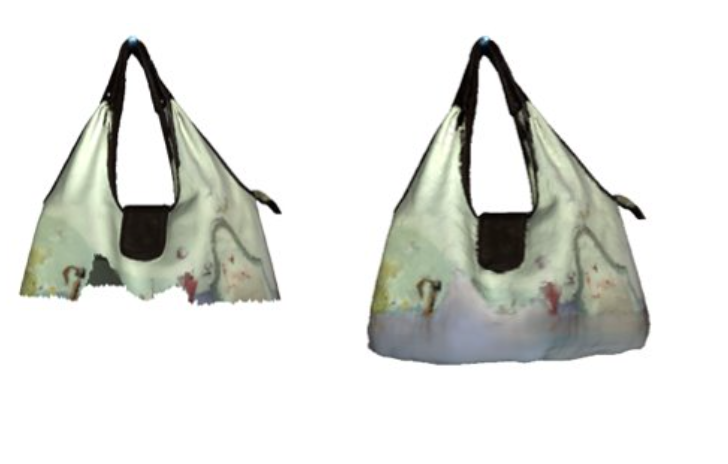}
     \end{subfigure}
     \begin{subfigure}[t]{0.2\textwidth}

         \includegraphics[width=\textwidth]{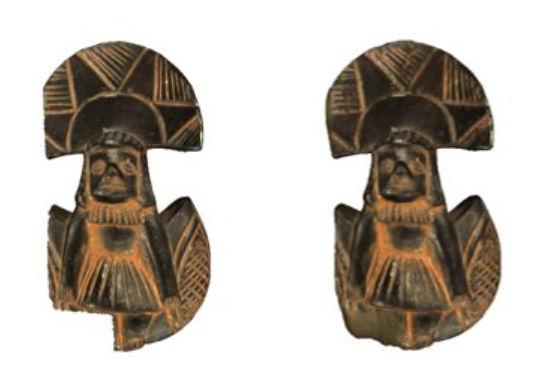}
     \end{subfigure}

     \begin{subfigure}[t]{0.2\textwidth}

         \includegraphics[width=\textwidth]{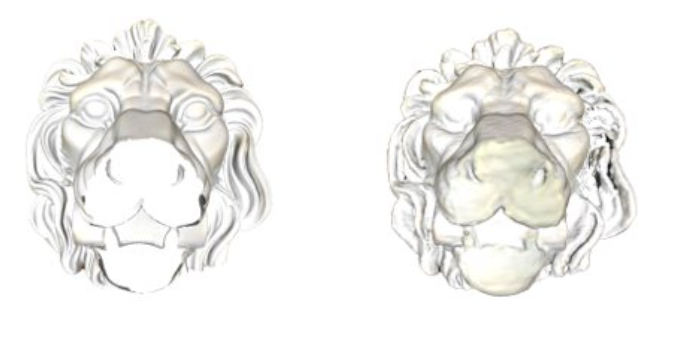}
     \end{subfigure}
      \begin{subfigure}[t]{0.2\textwidth}

         \includegraphics[width=\textwidth]{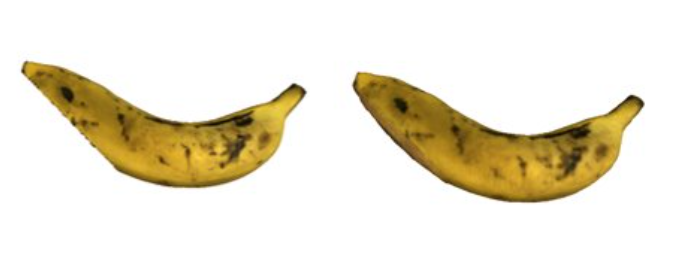}

     \end{subfigure}

        \caption{Visualization of  results in track 2. The left is partial input and the right is the reconstruction result. Our model can achieve great performance to recover the partial scans. Using the predicted bounding box as prior, the completion is coherent with shape and texture.}
        \label{fig:track2_result}
\end{figure}

The dataset consists of different generic objects with considerable scale variation in track 2. Compared to the model using the consistent bounding box, our method to predict possible bounding box as geometry prior for the specific object can generate more reasonable results. Some representative results are shown in Fig. \ref{fig:track2_result}. 

\subsection{Ablation Study}

\begin{table}[ht]
\centering
\begin{adjustbox}{width=0.35\textwidth}
\begin{tabular}{cccccccc}
\hline
IF-Net               & SMPL-fusion        &  BL    & \multicolumn{1}{c|}{SF} & final                      \\ \hline
\checkmark            &                      &                      & \multicolumn{1}{c|}{}             & 82.29                                \\
\checkmark            & \checkmark            &                      & \multicolumn{1}{c|}{}             & 83.26                               \\

\checkmark            & \checkmark            & \checkmark            & \multicolumn{1}{c|}{}             & 83.58                               \\
\checkmark            & \checkmark            & \checkmark            & \multicolumn{1}{c|}{\checkmark}    & \textbf{84.68}             \\ \hline
\multicolumn{1}{l}{} & \multicolumn{1}{l}{} &  \\

& \multicolumn{1}{l}{} & \multicolumn{1}{l}{} & \multicolumn{1}{l}{} & \multicolumn{1}{l}{}
\end{tabular}
\end{adjustbox}
\vspace{-8mm}
\caption{Ablation study of SHARP Challenge 1 Track 1. BL denotes balanced BCE loss and SF denotes shape fusion.}
\label{tab:ab1}
\end{table}
\begin{table}[ht]
\centering
\begin{adjustbox}{width=0.35\textwidth}
\centering
\begin{tabular}{cccccc}
\hline
IF-Net               & Bbox                 & \multicolumn{1}{c|}{Post-processing} & final          
\\ \hline
\checkmark            &                      & \multicolumn{1}{c|}{}             & 62.34               
\\
\checkmark            & \checkmark            & \multicolumn{1}{c|}{}             & 64.14                 
\\
\checkmark            & \checkmark            & \multicolumn{1}{c|}{\checkmark}    & \textbf{69.56}    
\\ \hline
\multicolumn{1}{l}{} 
\\
\multicolumn{1}{l}{} 
\end{tabular}
\end{adjustbox}
\vspace{-8mm}
\caption{Ablation study of SHARP Challenge 1 Track 2. Bbox denotes the completeness-aware bounding box prediction. Post-processing denotes shape fusion and completeness-based filter}
\label{tab:ab2}
\vspace{-3mm}
\end{table}
In this part, we remove some of the modules of the model to have a better understanding of how exactly our strategy improves the model performance.
In track 1, Table \ref{tab:ab1} shows the effect of SMPL model fusion, balanced loss and post-processing.The fusion of geometry prior from SMPL model can make great use of the human pose to obtain more accurate shape. The balanced loss can optimize the training procedure to reason about the missing parts. The post-processing can select better completion from output and input.
The result of Table \ref{tab:ab2} demonstrates that the predicted bounding box can address the variation of the scale effectively to get more accurate geometry. The post-processing strategy also improves the quality of the full mesh.
\section{Conclusions}
Our solution to the 3rd SHARP challenge improves existing approaches by incorporating learned geometric priors. To handle partial inputs with too many missing parts, we add priors of the general shape by a fusion module that combines a SMPL model generated from pose prediction with the input. To address varying appearances and sizes of the objects, we introduce completeness-aware bounding box adaptation to learn an appropriate prior for the magnitude of the recovered shape. Our future work includes improving the heuristic to reject bad predictions and finding better ways to incorporate the SMPL model.



{\small
\bibliographystyle{ieee}
\bibliography{egbib}
}

\end{document}